# Predicting Consumer Purchasing Decisions in the Online Food Delivery Industry

Batool Madani and Hussam Alshraideh

Department of Industrial Engineering,
American University of Sharjah, Sharjah, UAE

## Abstract

*This transformation of food delivery businesses to online platforms has gained high attention in recent years. This due to the availability of customizing ordering experiences, easy payment methods, fast delivery, and others. The competition between online food delivery providers has intensified to attain a wider range of customers. Hence, they should have a better understanding of their customers' needs and predict their purchasing decisions. Machine learning has a significant impact on companies' bottom line. They are used to construct models and strategies in industries that rely on big data and need a system to evaluate it fast and effectively. Predictive modeling is a type of machine learning that uses various regression algorithms, analytics, and statistics to estimate the probability of an occurrence. The incorporation of predictive models helpsonline food delivery providers to understand their customers. In this study, a dataset collected from 388 consumers in Bangalore, India was provided to predict their purchasing decisions. Four prediction models are considered: CART and C4.5 decision trees, random forest, and rule-based classifiers, and their accuracies in providing the correct class label are evaluated. The findings show that all models perform similarly, but the C4.5 outperforms them all with an accuracy of 91.67%.*

## Keywords

*Food Delivery Industry, Purchasing Prediction, Machine Learning, Decision Trees, Random Forest, Rule-Based Classifier.*

## 1. Introduction

The internet's fast evolving technology has infiltrated practically every part of our lives, giving limitless opportunity for businesses and customer relationships. It has boosted online food services by allowing consumers to search, compare the provided services, and select companies with high service performance. The food delivery industry has experienced a lot of transformations, moving from restaurant-to-consumer delivery to platform-to-consumer delivery [1]. At first, restaurants had dedicated websites or phone numbers to allow customers to place their orders as well as have a dedicated delivery team. Then, in 2013, the business of platform-to-consumer came, in which specialized delivery players provide logistic support for restaurants. This new business model, which is referred to as online food delivery, provides customers with the abilityto compare menus, check reviews, and place many orders from different restaurants at the same time. Online food delivery has grown by 25% between 2015 and 2018 and it is expected to grow at a rate of 10.7% by 2023 [2]. Online food delivery is defined as the process of food online ordering, order process and preparation, and delivery. The platforms of online food delivery such as Uber Eats, Deliveroo, and Zomato provide a variety of functions, such as providing customers with a wide variety of food choices, taking the order and transferring it to the food provider, monitoring the payment and managing the delivery of the food [1]. Worldwide, China leads the way in the market share of this industry, followed by the US and then India. Between 2020 and 2024, the market's revenue is expected to grow at a 7.5% annual rate, resulting in a market volume of US $182,327 million by 2024 [3]. This indicates the rapid increase in the online food delivery market, which intensifies the competition between companies to gain dominance among others and increases the need to determine the key success factors that are critical to online food delivery providers. Online food delivery providers need to gain insights and reviews from customers to capture a larger segment of the market share. Besides, the decision ofwhether or not to make a purchase is a complex process, impacted by many aspects.

Due to the major dependence of consumers on online services, the online food delivery industry has been proactive in a way thatcreates a highly competitive market, which makes companies more susceptible to losing their marketplace. In order for companies to gain a competitive edge over others, they should understand their customers' needs, expectations, and requirements. Otherwise, the misunderstanding of customers' expectations can lead to the loss of customers' purchases and commitment. The reasoning behind that requires the evaluation of multiple measures, such as the timing of deliveries, the performance of tracking systems, quality of packaging, food temperature, food freshness, etc. Because obtaining customers' purchasing decisions is critical, determining the right reasoning can increase customer satisfaction and thus improve the company's market position. This requires the collection of customer data about their evaluation of the performance of online food delivery companies in terms of different aspects. While purchase prediction has been discussed in consumer research for a long time, the emergence of customer analytics has reignited such issues recently. One possible way of exploiting the data of customers' purchasing decisions is via machine learning techniques to construct accurate prediction models. Machine learning is a highly advanced, rapid, and accurate technology[4]. In the customer relationship management domain, the use of machine learning techniques for predictive purposes on a customer base is frequently investigated, with customer churn prediction being the most prominent goal. For maintaining customer relationships, accurate prediction of a customer's activity state and future purchasing propensities are critical [5]. Predicting purchasing decisions is a time series forecasting task that can be solved using traditional statistical techniques such as autoregressive moving average [6]. However, machine learning techniques are often more powerful and versatile, when dealing with time series forecasting. This is because they enable the employment of cutting-edge supervised learning algorithms like regression support vector machines and model trees.

In this study, machine learning techniques are used to anticipate customer purchasing decisions in the context of online food delivery. This is accomplished through the use of a dataset about customer purchasing experiences, which covers a variety of characteristics related to online food delivery providers. A comparison of three prediction models will be provided in order to determine which model is the most suitable and provides the best performance in terms of accuracy. The remainderof the paper is structured as follows. In section 2, the literature review is presented, while section 3 describes the used dataset. This is followed by the analysis and the results in section 4. Finally, section 5 provides the conclusion.

## 2. LITERATURE REVIEW

The rapid increase in demand for online services has motivated practitioners and academicians to seek a better understanding of customers' purchasing decisions and behaviors. An increasing number of studies are adopting prediction models to forecast purchasing decisions under different problem settings and varied inputs [7-16]. Van Den Poel and Buckinx [7], investigated the impact of different sets of predictors on online purchasing behavior using logit modeling. The logit modeling method is used to answer the question of whether or not a purchase will be made during the next visit using a set of predictors: general clickstream behavior, detailed clickstream behavior, customer demographics, and historical purchase behavior. Using the same prediction model, Yilmaz and Belbag [8] predicted consumer behavior regarding purchasing remanufactured products, which indicated that low prices, product reliability, and product promotions affect positively the purchasing decisions of consumers. In the context of product promotions, Ling et al. [9] proposed a feature-combined deep learning framework for predicting consumers' purchase intent during promotions across multiple online channels. The study also suggested that including demographics information enhances the prediction performance, but, increases the methodological challenge. The importance of demographic information has also been considered in [10], which emphasized the importance of defining the demographic of people in a certain region to the marketing of the automobile industry in order to define the target group and integrate marketing strategies to enhance the purchase decision of a car. Therefore, they have investigated the prediction of consumer purchase decisions using the demographic structure of premium car owners using the logistic regression classification model. Due to the complexity of online marketsand the diversity of their consumers, prediction models with powerful self-learning capabilities, such as artificial neural networks, decision trees, and random forest, to name a few, are increasingly relied on. Gupta and Pathak [11] applied different classification algorithms, such as decision trees, support vector machine, and rule-based method, to predict customers' purchase decisions, whether a user will be interested in buying a certain set of products that are placed in the online shopping cart or not. Similarly, Tang et al. [12] developed a hybrid model that is based on the technique of support vector machine and the firefly algorithm, for predicting online-purchasing behavior to forecast whether or not a customer will purchase during the next visit. Martínez et al. [13] developed an advanced analytics technique for non-contractual customer behavior prediction by establishing a dynamic and data-driven machine learning framework. Among the state-of-the-art machine learning algorithms, the gradient tree boosting method has outperformed the other methods and provided a prediction accuracy of 89%. Liao and Tsai [14] proposed a multimodel fusion B2C online marketing algorithm based on the least squares-support vector machine method, which has proved to have a high prediction accuracy compared to traditional prediction single-model.

Wang and Xu [15] examined the Chinese government's introduction of a 7-day unreasonable return policy to boost customer trust in e-commerce companies. The ease of return has a direct impact on customer purchase decisions, which is investigated in this study. An ensemble learning method based on a fuzzy support vector machine is used to predict customers' purchasing intentions. The proposed method outperformed a set of several classifiers such as logistic regression, support vector machine, and random forest in terms of prediction accuracy. Ghosh and Banerjee [16] proposed a modified random forest algorithm-based predictive analytic methodology. Using five parameters (previous purchasing habits, a sequence of online advertisements viewed, customer location, number of clicks, and last used service), the model seeks to predict purchasing decisions in cloud services. The model also had high forecast accuracy, with online advertisements being the most important component in making a purchase decision.

In the context of online food delivery, Natarajan et al. [17] investigated the impact of online food delivery service providers such as Swiggy, Foodpanda, and Zomato on Indian consumer preferences in the setting of online food delivery. According to the study's findings, consumers favor originality in terms of pricing, quality, and delivery. The online food delivery market in India is one of the world's largest markets. According to a study conducted in the years, 2019-2020 [18], the Indian online food delivery market was estimated to be valued at $4.35 billion in 2020. This was a significant gain over the previous year, when the market was estimated to be

worth roughly 2.9 billion dollars. In addition, the food delivery sector is predicted to reach about 13 billion dollars in value by 2025. According to Anusha and Panda [19], "young India's appetite is one of the key drivers for demand in the food and beverage industry on the whole". As a result, the analysis provided in this study is centered on the consumers of India. Furthermore, compared to other fields of prediction research, existing research on online purchasing choice prediction is limited, particularly in the application of online food delivery. Therefore, the purchasing decisionsin an online food delivery segment will be investigated here, which will help decision makers anticipate their customers' buying intentions and determine the most influential factors in purchasing decisions.

## 3. DATA DESCRIPTION

To predict whether the consumer will buy again or not, a dataset obtained from the open-source database Kaggle is used [20]. The obtained dataset was collected from 388 consumers in Bangalore, India, and it has 55 variables consisting of the consumers' demographics and consumers' inputs about the delivery service, including the time, packaging, delivery person, and many others. There are 25 variables with a 5-point Likert-type scale (1 = Strongly disagree, 5 = Strongly agree), 8 variables about the level of importance of certain aspects, 10 demographic variables, 2 categorical variables with three levels, about the influence of delivery timing and the rating of restaurants, and a combination of categorical and numerical variables. Finally, the response variable is a categorical variable with two classes: "will purchase (yes)" and "will not purchase (no)".

## 4. RESULTS AND ANALYSIS

As a first step, data pre-processing will be carried out, in which some of the input variables will be eliminated. Secondly, an exploratory data analysis will be performed to summarize the data, obtain insights and understanding of the demographics of the consumers, and investigate the relationship between the purchase decision and the other attributes. Due to a large number of input attributes, feature selection and elimination are considered to reduce the number of inputs and determine the significant ones. Table 1 provides a detailed description of the attributes and the variables of interest. Finally, using the significant attributes, different classification methods, which are decision-tree, random forest, and rule-based classifier, are used to predict the purchase decision. A comparison between them will be made based on the accuracy and the significance of the difference between them. The classification models will be first exposed to model parameter tuning using cross-validation to enhance their performance, and then will be tested on a new dataset to evaluate their performance.

### 4.1. Exploratory Data Analysis

Prior to this analysis, data preprocessing is performed, in which some of the variables are removed due to their irrelevancy to the problem, such as latitude, longitude, pin code, reviews. Afterward, an analysis of the demographics and the preferences of participants is presented. Table 2 presents the demographics summary of the participants.

As it can be observed, the mean average age of respondents is 25 years. The mix of respondentsis fairly balanced, with males contributing to 57.2%. In terms of marital status, singles (69.1%) have a comparatively large presence, followed by married. Most of the respondents were students (53.3%), followed by employees (30.4%). For educational qualifications, graduates (45.6%) followed by postgraduates (44.9%) represent the majority of the respondents. Additionally, the majority of the respondents (46.3%) live with 3-4 members.

Table 1. Dataset description

| | Attribute | Type | Description |
|---|---|---|---|
| 1 | Age | Integer | Age of participants |
| 2 | Gender | Character | Gender of participants |
| 3 | Marital Status | Character | Marital status of participants |
| 4 | Occupation | Character | Job occupation of participants |
| 5 | Monthly income | Character | Monthly income of participants |
| 6 | Educational Qualifications | Character | Educational qualification of participants |
| 7 | Family size | Integer | Number of family members/ friends living with |
| 8 | Ordering medium preference 1 | Character | Through which medium participants are ordering |
| 9 | Ordering medium preference 2 | Character | Through which medium participants are ordering |
| 10 | Meal preference 1 | Character | What type of meal participants are ordering |
| 11 | Meal preference 2 | Character | What type of meal participants are ordering |
| 12 | Ordering ease and convenience | Character | Ease and convenience of online ordering |
| 13 | Time saving | Character | Does it save time? |
| 14 | Restaurant choices | Character | More restaurant choice influence |
| 15 | Easy payment option | Character | Payment option influence |
| 16 | More offers and discounts | Character | Offers and discount influence |
| 17 | Good food quality | Character | Food quality influence |
| 18 | Good tracking system | Character | Tracking system influence |
| 19 | Self-cooking | Character | Self-cooking causes not purchasing |
| 20 | Health concern | Character | Health concern causes not purchasing |
| 21 | Late delivery | Character | Later Delivery causes not purchasing |
| 22 | Poor hygiene | Character | Poor Hygiene causes not purchasing |
| 23 | Bad experience | Character | Past experiences cause not purchasing |
| 24 | Unavailability | Character | Unavailability causes not purchasing |
| 25 | Unaffordable | Character | Un-affordability causes not purchasing |
| 26 | Long delivery time | Character | Long delivery causes cancellation |
| 27 | Delay of delivery person | Character | Delay of delivery person assigned causes cancellation |
| 28 | Delay of picking up food | Character | Delay of delivery person picking up food causes cancellation |
| 29 | Wrong order delivered | Character | Previous wrong order causes cancellation |
| 30 | Missing item | Character | Missing item in order causes cancellation |
| 31 | Order placed by mistake | Character | Placed order by mistake causes cancellation |
| 32 | Influence of delivery time | Character | Time of delivery influencing purchasing decision |
| 33 | Order time | Character | When do you order? |
| 34 | Maximum waiting time | Character | How long can you wait? |
| 35 | Residence in busy locations | Character | Residence in busy location |
| 36 | Google maps accuracy | Character | My location in google maps is accurate |
| 37 | Good road conditions | Character | My residence area road condition is good |
| 38 | Low quantity | Character | low quantity low delivery time |
| 39 | Delivery person ability | Character | Delivery person ability depends on time of delivery |
| 40 | Influence of restaurant rating | Character | Rating of restaurant influencing purchasing decision |
| 41 | Less delivery time | Character | Importance of Less delivery time |
| 42 | High quality of package | Character | Importance of Quality of package |
| 43 | Number of calls | Character | Importance of Number of calls made by delivery captain |
| 44 | Politeness | Character | Importance of Politeness of delivery captain |
| 45 | Freshness | Character | Importance of Freshness of food |
| 46 | Temperature | Character | Importance of Temperature of food |
| 47 | Good taste | Character | Importance of taste |
| 48 | Good quantity | Character | Importance of Quantity in food |
| 49 | Purchasing decision | Character | Will the customer purchase again (output variable) |

Table 2. Demographics Summary

| Category | Subcategory | Value |
|---|---|---|
| Age | Mean | 24.6 years |
| Gender | Female | 42.8% |
| | Male | 57.2% |
| Marital Status | Married | 27.8% |
| | Prefer not to say | 3.1% |
| | Single | 69.1% |
| Occupation | Employee | 30.4% |
| | Housewife | 2.3% |
| | Student | 53.3% |
| | Self Employed | 13.9% |
| Educational qualifications | Graduate | 45.6% |
| | Ph.D | 5.9% |
| | Postgraduate | 44.9% |
| | School | 3.1% |
| | Uneducated | 0.5% |
| Family size | Less than 3 | 32.2% |
| | 3-4 | 46.3% |
| | 5-6 | 21.4% |

Table 3 shows that customers prefer to use food delivery applications the most, ordering mostly food for snacks (32.0%) and dinner (80.4%).

Table 3. Preference Summary

| Category | Subcategory | Percentage |
|---|---|---|
| Ordering medium preference 1 | Direct call | 1.3% |
| | Food delivery apps | 92.3% |
| | Walk-in | 5.7% |
| | Web browser | 0.8% |
| Ordering medium preference 2 | Direct call | 53.6% |
| | Walk-in | 26.8% |
| | Web browser | 19.6% |
| Meal preference 1 | Breakfast | 13.7% |
| | Dinner | 23.4% |
| | Lunch | 30.9% |
| | Snacks | 32.0% |
| Meal preference 2 | Dinner | 80.4% |
| | Lunch | 7.2% |
| | Snacks | 12.4% |
| Cuisine preference 1 | Bakery items (snacks) | 0.3% |
| | Non-Veg foods (Lunch / Dinner) | 81.2% |
| | Sweets | 0.8% |
| | Veg foods (Breakfast / Lunch) | 17.8% |
| Cuisine preference 2 | Bakery items (snacks) | 3.4% |
| | Ice cream / Cool drinks | 9.0% |
| | Sweets | 11.9% |
| | Veg foods (Breakfast / Lunch) | 75.8% |

Figure 1 summarizes the relationship between purchase decisions and gender and marital status, respectively. Single customers are more likely to use online food delivery services than customers who are not single. Males and females use online services in comparable amounts, with males having a higher proclivity to buy from online food providers.

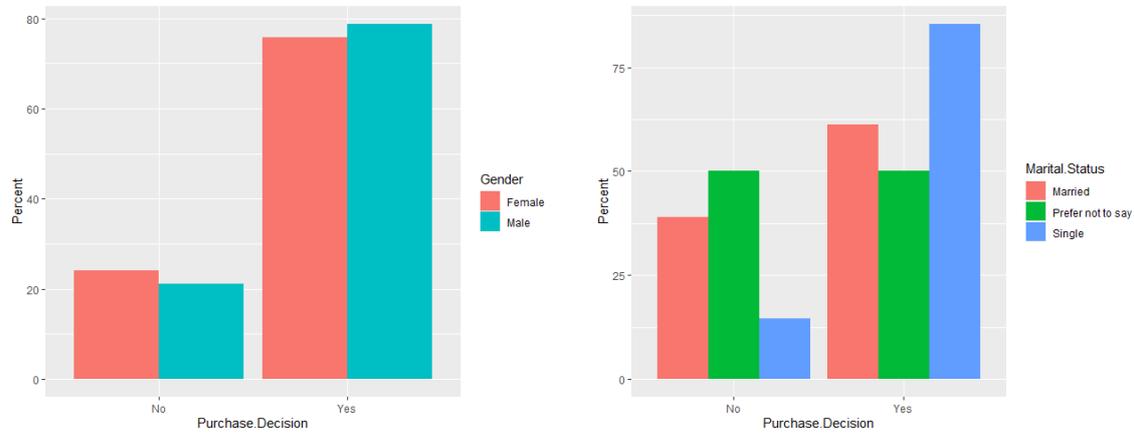

Figure 1. Relationship between purchasing decision and gender and marital status

## 4.2. Prediction Models

Three prediction models: decision tree, random forest, and rule-based classifier, will be compared based on their performances. In all three prediction models, the attributes of marital status, occupation, educational qualifications, family size, ordering medium preferences, meal preferences, and cuisine preferences are eliminated. In all three models, the training data covers 75% of the data, while the remaining 25% is assigned to test data. The method for data training is cross-validation with 10 folds. Finally, they are compared based on prediction accuracy. The prediction accuracy is derived from the confusion matrix which summarizes a classifier's classification performance in relation to some test data. It's a two-dimensional matrix with the true class of an object in one dimension and the class that the classifier assigns in the other [21]. The confusion matrix is frequently used with two classes, one of which is labeled as positive and the other as negative. True positives (TP), false positives (FP), true negatives (TN), and false negatives (FN) are the four cells of the matrix in this context (FN). The following presents a description of each parameter [4].

- True positive (TP): A positive sample predicted by the model.
- False positive (FP): A negative sample predicted by the model as a positive example.
- False negative (FN): The positive sample predicted by the model is used as a negative sample.
- True negative (TN): A sample predicted to be negative by the model.

The prediction accuracy is defined as the number of correct predictions divided by the total number of input samples. It is calculated as the following:

$$Accuracy = (TP + TN) / (TP + TN + FP + FN)$$

In this study, all the calculations of confusion matrices, prediction accuracies, and other parameters are performed using R software.

**4.2.1. Decision tree**

**Classification and regression tree (CART)**: The CART decision tree is used for regression predictive modeling problems. It is a binary recursive partitioning tree, where each parent node in the tree is split into two child nodes [22]. Further, CART is known for its simple interpretationand inherent logic. Here, CART is used to predict the purchasing decisions of online food consumers. From Figure 2, we can obtain the following conclusions:

- The probability of a customer purchasing the next time, who evaluated the "ease and convenience of ordering" and "ordering saving time" elements with more than 3 is (1-0.06) = 0.94, and this node covers 77% of the dataset.
- The probability of a customer who will not purchase the next time, who evaluated the ease and convenience element with less than 3 is (1-0.85) = 0.15, and this node covers 18% of the dataset.

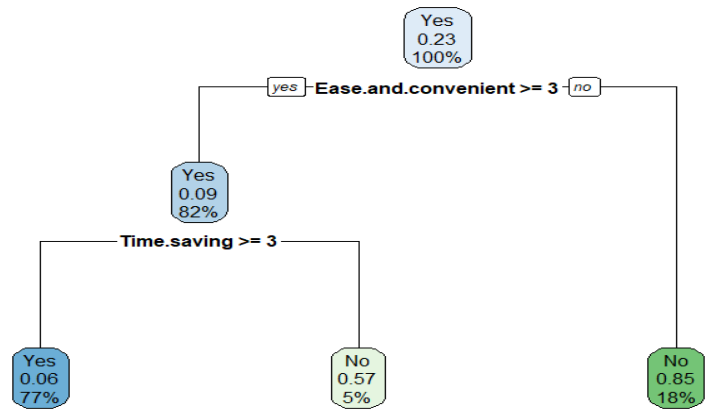

Figure 2. CART decision tree

For training data, the confusion matrix shows that 71.6% of those who will purchase again are classified correctly and 15.4% of those who will not purchase again are classified correctly. On the other hand, 5.8% of consumers are classified as not purchasing, while 7.2% of consumers are classified as purchasing wrongly. The accuracy of the CART tree is 87%. On other hand, the accuracy of this tree on the test data is 84.38%, which implies that CART is performing well.

**C4.5 decision tree**: In Data Mining, the C4.5 algorithm is utilized as a decision tree classifier, which can be used to make a decision based on a sample of data [23]. It is known for its ability to work with discrete and continuous data as well as handling incomplete data. After implementing the C4.5 tree, its accuracy on both training and testing data outperforms the CART decision tree, as seen in Figures 3 and 4.

```
          Reference
Prediction  Yes    No
       Yes  71.6   7.2
       No    5.8  15.4

Accuracy (average) : 0.8699
```

```
          Reference
Prediction  Yes    No
       Yes  74.7   6.2
       No    2.8  16.2

Accuracy (average) : 0.9098
```

Figure 3. Confusion matrices based on training data (CART – C4.5 decision trees)

```
              Reference
Prediction Yes No
       Yes  66  6
       No    9 15

               Accuracy : 0.8438
                 95% CI : (0.7554, 0.9098)
    No Information Rate : 0.7812
    P-Value [Acc > NIR] : 0.08356

                  Kappa : 0.5652

 Mcnemar's Test P-Value : 0.60558

            Sensitivity : 0.8800
            Specificity : 0.7143
         Pos Pred Value : 0.9167
         Neg Pred Value : 0.6250
             Prevalence : 0.7812
         Detection Rate : 0.6875
   Detection Prevalence : 0.7500
      Balanced Accuracy : 0.7971

       'Positive' Class : Yes
```

```
              Reference
Prediction Yes No
       Yes  73  6
       No    2 15

               Accuracy : 0.9167
                 95% CI : (0.8424, 0.9633)
    No Information Rate : 0.7812
    P-Value [Acc > NIR] : 0.0003671

                  Kappa : 0.7382

 Mcnemar's Test P-Value : 0.2888444

            Sensitivity : 0.9733
            Specificity : 0.7143
         Pos Pred Value : 0.9241
         Neg Pred Value : 0.8824
             Prevalence : 0.7812
         Detection Rate : 0.7604
   Detection Prevalence : 0.8229
      Balanced Accuracy : 0.8438

       'Positive' Class : Yes
```

Figure 4. Confusion matrices based on testing data (CART – C4.5 decision trees)

The resulted C4.5 tree is illustrated in Figure 5, and the following conclusions are obtained.

- If the "ease and convenient" and "good taste" are rated with less than or equal to 2, the customer will purchase again with a probability of 100%.
- If the "ease and convenient" is rated with less than or equal to 2 and "good taste" was given a rate of greater than 2, the customer will not purchase again with a probability of 90%.
- If the "ease and convenient" and "time saving" were given a rate of greater than 2, the customer will purchase again with a probability of 95%.
- If the "ease and convenient" was given a rate of more than 2, "time saving" was given a rate of less or equal to 2, and "more offers and discounts" was rated greater than 4, the customer will purchase again with a probability of 100%.
- If the "ease and convenient" was given a rate of more than 2, "time saving" was given a rate of less or equal to 2, "more offers and discounts" was rated less or equal to 4, and the age of the consumer is greater than 25, the customer will not purchase again with a probability of 100%.
- If the "ease and convenient" was given a rate of more than 2, "time saving" was given a rate of less or equal to 2, "more offers and discounts" was rated less or equal to 4, age of the consumer is less than/equal to 25, and there is no influence of restaurant rating, the customer will not purchase again with a probability of 100%.
- If the "ease and convenient" was given a rate of more than 2, "time saving" was given a rate of less or equal to 2, "more offers and discounts" was rated less or equal to 4, age of the consumer is less than/equal to 25, there is an influence of restaurant rating, and the rate of "good road condition" is greater than 2, the customer will purchase again with a probability of 100%.
- If the "ease and convenient" was given a rate of more than 2, "time saving" was given a rate of less or equal to 2, "more offers and discounts" was rated less or equal to 4, age of the consumer is less than/equal to 25, there is no influence of restaurant rating, and the rate of good road condition is less than/equal to 2, the customer will not purchase again with a probability of 100%.

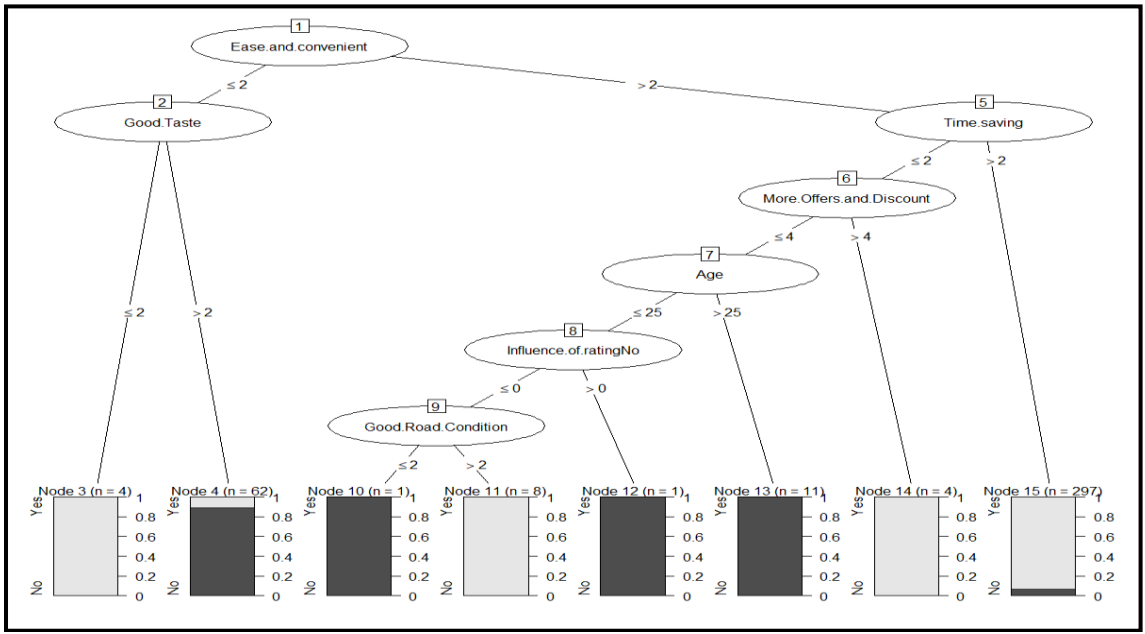

Figure 5. C4.5 decision tree

In other words, the predictors selected by the C4.5 decision tree are the importance of good taste, age, the influence of restaurant rating, ordering ease and convenience, the goodness of road conditions, time-saving, and availability of offers and discounts.

#### 4.2.2. Random forest

A random forest is made up of many separate decision trees that work together to form an ensemble. Each tree in the random forest produces a class prediction, and the class performing the best becomes the prediction of the model [24]. When applied to our problem, it outperforms the C4.5 decision tree on training, with a 94.18% accuracy rate. On the other hand, its performance on the test is comparable to that of the C4.5 decision tree (90.62%).

#### 4.2.3. Rule-based classifier

The rule-based classifier is employed in the class prediction method to give the rules a rating, which is then used to predict the class of future cases [25]. When compared with the other prediction models used, it performs less than the random forest model, in which its accuracy on the training data is 91.44% while on the testing data is 87.5%. Figures 6 and 7 show the comparison between the random forest and the rule-based classifier. The resulted rule-based classifier, shown in Figure 8, draws the following findings.

- If the "ease and convenience" and "time saving" are rated above 2 and "unaffordable" is rated less or equal to 3, the consumer will decide to purchase.
- If the "ease and convenience" is rated less or equal to 3 and "low quantity-time" are rated above 1, and female, the consumer will not decide to purchase.
- If the "ease and convenience" is given a rate of greater than 3. "More restaurant choices" is rated above 2, "good tracking system" is rated less than/equal to 3, a consumer is a female, and the "delay of the delivery person assigned" was rated less than/equal to 4, the consumer will decide to purchase.
- If the consumer is a female with age less than/equal to 30, who gave a rate above 2 for "more restaurant choices", he/she will purchase again.
- If the rate of "low quantity-low time" is above 1, "time saving" is less than/equal to 4, "ease and convenient" is greater than 1. And "wrong order delivered" is less than/equal to 4, the consumer will not decide to purchase again.
- If there is an influence of the time, and order time is not on Saturday or Sunday, and "late delivery" was given a rate of less than/equal to 4, the consumer will purchase again.
- If "self-cooking" is given a rate above 3, the consumer will not purchase again.
- If the age of the consumer is less or equal to 25 years, then he/she is expected to purchase again.
- If everything other than the aforementioned is not satisfied, the consumer will not purchase again.

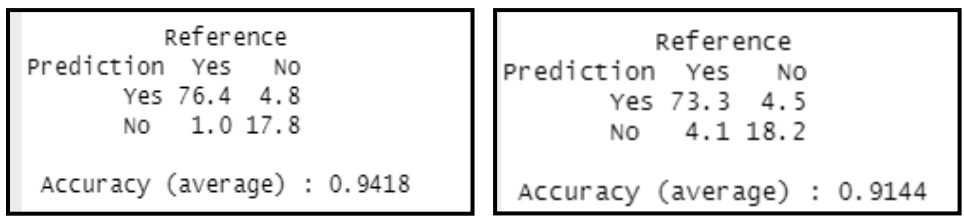

Figure 6. Confusion matrices based on training data (random forest – rule-based classifier)

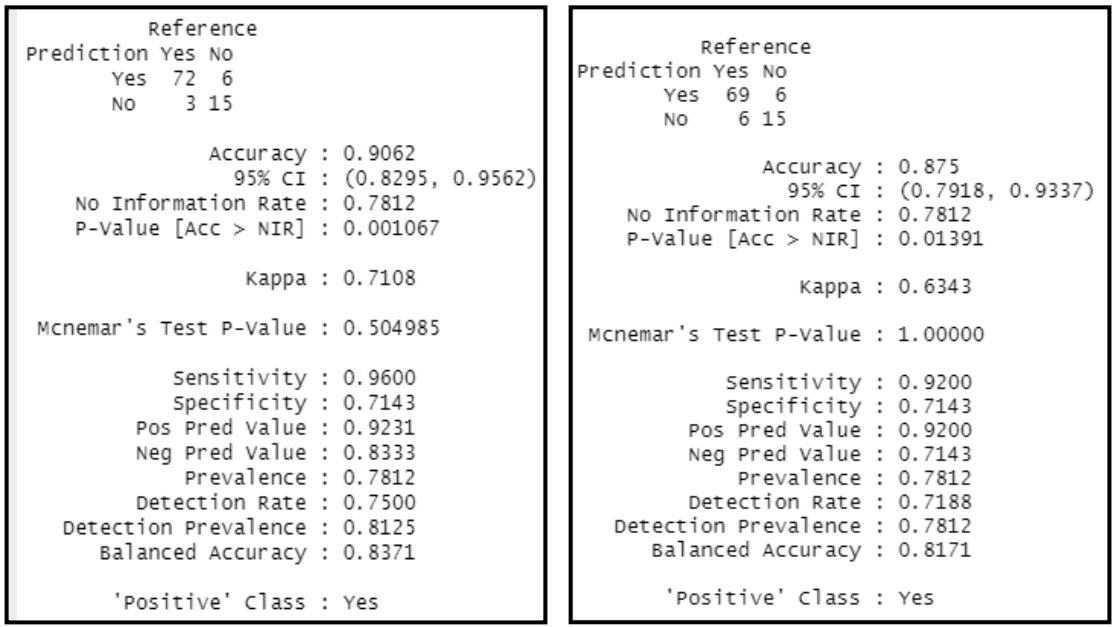

Figure 7. Confusion matrices based on testing data (random forest – rule-based classifier)

### 4.2.4. Comparison

As seen in the previous section, there is a difference in the accuracies of the four classification models. However, this difference should be evaluated based on the p-value criterion to determine if the difference is significant. Figure 9 shows that based on accuracy, there is a significant difference between the CART decision tree and therule-based classifier, and the random forest.

Add to that, based on the conclusions obtained previously, the random forest and C4.5 models perform comparably. However, the accuracy provided by the C4.5 decision is better.

```
PART decision list
------------------

Ease.and.convenient > 2 AND
Time.saving > 2 AND
Unaffordable <= 3: Yes (185.0/4.0)

Ease.and.convenient <= 3 AND
Low.quantity.low.time > 1 AND
Ease.and.convenient > 1 AND
GenderMale <= 0: No (26.0)

Ease.and.convenient > 3 AND
More.restaurant.choices > 2 AND
Good.Tracking.system <= 4 AND
GenderMale > 0 AND
Delay.of.delivery.person.getting.assigned <= 4: Yes (13.0)

GenderMale <= 0 AND
Age <= 30 AND
More.restaurant.choices > 2: Yes (17.0)

Low.quantity.low.time > 1 AND
Time.saving <= 4 AND
Ease.and.convenient > 1 AND
wrong.order.delivered <= 4: No (30.0/1.0)

Influence.of.timeYes > 0 AND
Order.TimeWeekend (Sat & Sun) <= 0 AND
Late.Delivery <= 4: Yes (12.0)

Self.Cooking > 3: No (5.0)

Age <= 25: Yes (2.0)

: No (2.0)

Number of Rules  :       9
```

Figure 8. Rule-based classifier

```
Accuracy
             PART      RandomForest CART       C4.5
PART                   -0.027135    0.045304   0.004845
RandomForest 0.77657                0.072438   0.031980
CART         0.03673   0.02140                 -0.040459
C4.5         1.00000   0.37712      0.42723

Kappa
             PART      RandomForest CART       C4.5
PART                   -0.06748     0.14927    0.02931
RandomForest 1.00000                0.21675    0.09679
CART         0.11409   0.07706                 -0.11997
C4.5         1.00000   0.42192      0.79425
```

Figure 9. Comparison between the classification models based on p-value

## 5. CONCLUSION

In this study, we used several prediction models to determine whether a customer would purchase again from the online food delivery platforms. The ability to do so provides a strong predictive tool for online food delivery providers to have a better understanding of their customers, and to improve their services accordingly. Building the right prediction mode, which combines high prediction accuracy with sound reasoning, can assist decision-makers in reaching accurate conclusions about the major determinants of customer satisfaction, hence increasing the likelihood of repeat purchases. Past research has considered the implementation of prediction models on purchasing decisions. However, a limited number of studies have incorporated their use into the industry of online food delivery. In this study, we used CART and C4.5 decision trees, a random forest, and a rule-based classifier. The four models performed outstandingly in predicting the purchasing decision, but the C4.5 decision tree performed the best, by providing an accuracy of 91.67%.

Among other algorithms, the C4.5 algorithm is a decision tree algorithm that can be used to build rules that are easy to understand and fast. The approach can also provide a basic model subsystem that can be utilized to support a decision-making system. The C4.5 decision tree has an improved tree pruning strategy that lowers misclassification errors in the training data set owing to noise and too much information. Add to that, it can handle missing attributevalues as well as handling different types of data. However, it is only used for small datasets where all or a fraction of the entire dataset must be kept in memory permanently. As a result, its suitability for mining massive databases must be examined. In addition, an improved version of the traditional prediction models must be developed to enhance their accuracy and the time taken to derive the tree. The pruning strategy of C4.5 may allow the trimming of nodes with high value information. Thus, adding enhancements and treatments to the selection of the nodes to be trimmed can increase the output accuracy.

## AUTHORS

**Batool Madani** holds a B.Sc. in Nuclear Engineering. She received her M.Sc. in Engineering Systems Management from the American University of Sharjah, U.A.E, in 2019. She is currently a PhD candidate and a Graduate Teaching Assistant at the American University of Sharjah. Her research is oriented around the integration of technologies in the Last Mile delivery problem. Her research interests include wireless technologies, machine learning, decision making, drones, logistics, and optimization.

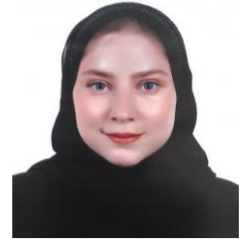

**Hussam Alshraideh** is an Associate Professor of Operations Research and Statistics at the Industrial Engineering Department at the American University of Sharjah (AUS). He holds a dual Ph.D. degree in Industrial Engineering and Operations Research with a minor in Statistics from The Pennsylvania State University. He also holds a master's degree in Industrial Engineering/Quality Engineering from Arizona State University. His current research interests include statistical process optimization and smart data analytics applications in healthcare related fields. He has published more than forty papers in highly reputable journals on health informatics and process control.

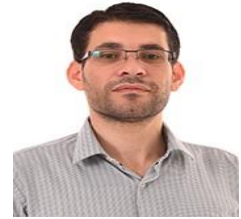